\documentclass[10pt,twocolumn,letterpaper]{article}

\usepackage{iccv}
\usepackage{times}
\usepackage{epsfig}
\usepackage{graphicx}
\usepackage{amsmath}
\usepackage{amssymb}

\usepackage{subcaption}
\usepackage{booktabs}
\usepackage{multirow}

\usepackage{caption} 
\usepackage{color} 
\usepackage[pagebackref=true,breaklinks=true,letterpaper=true,colorlinks,bookmarks=false]{hyperref}

\iccvfinalcopy 

\def\iccvPaperID{9952} 

\ificcvfinal\fi

\begin{document}

\title{AugNet: End-to-End Unsupervised Visual Representation Learning with Image Augmentation}

\author{
  Mingxiang Chen \\
  Beike.com \\
  Beijing, China 100085 \\
  {\tt\small chenmingxiang002@ke.com}
  
  \and
   
  Zhanguo Chang \\
  Beike.com \\
  Beijing, China 100085 \\
  {\tt\small changzhanguo001@ke.com}
  
  \and
   
  Haonan Lu \\
  Beike.com \\
  Beijing, China 100085 \\
  {\tt\small luhaonan001@ke.com}
  
  \and
   
  Bitao Yang \\
  Beike.com \\
  Beijing, China 100085 \\
  {\tt\small yangbitao001@ke.com}
  
  \and
  
  Zhuang Li \\
  Beike.com \\
  Beijing, China 100085 \\
  {\tt\small lizhuangneu@163.com}
  
  \and
  
  Liufang Guo \\
  Beike.com \\
  Beijing, China 100085 \\
  {\tt\small guoliufang001@ke.com}
  
  \and
   
  Zhecheng Wang \\
  Stanford University \\
  Stanford, CA 94305 \\
  {\tt\small zhecheng@stanford.edu}
}

\maketitle
\ificcvfinal\thispagestyle{empty}\fi

\newcommand{\zhecheng}[1]{{\textcolor{blue}{#1}}}
\newcommand{\question}[1]{{\textcolor{red}{#1}}}

\begin{abstract}
Most of the achievements in artificial intelligence so far were accomplished by supervised learning which requires numerous annotated training data and thus costs innumerable manpower for labeling. Unsupervised learning is one of the effective solutions to overcome such difficulties. In our work, we propose AugNet, a new deep learning training paradigm to learn image features from a collection of unlabeled pictures.
We develop a method to construct the similarities between pictures as distance metrics in the embedding space by leveraging the inter-correlation between augmented versions of samples. Our experiments demonstrate that the method is able to represent the image in low dimensional space and performs competitively in downstream tasks such as image classification and image similarity comparison. Specifically, we achieved over 60\% and 27\% accuracy on the STL10 and CIFAR100 datasets with unsupervised clustering, respectively. Moreover, unlike many deep-learning-based image retrieval algorithms, our approach does not require access to external annotated datasets to train the feature extractor, but still shows comparable or even better feature representation ability and easy-to-use characteristics. In our evaluations, the method outperforms all the state-of-the-art image retrieval algorithms on some out-of-domain image datasets.
The code for model implementation is available at https://github.com/chenmingxiang110/AugNet.
\end{abstract}

\section{Introduction}
Over the past few years, deep neural network has already shown its great capability and potential in the field of computer vision. From AlexNet \cite{krizhevsky2012imagenet} to Inception \cite{szegedy2016rethinking}, from RCNN \cite{girshick2014rich} to YOLO \cite{bochkovskiy2020yolov4}, from UNet \cite{ronneberger2015u} to DeepLab \cite{chen2017rethinking}, the footprints of deep learning algorithms have spread to many places where we never imagined to be able to stand. After ResNet \cite{he2016deep} surpassed humans’ average level in image classification tasks in 2015, the scale of victory has gradually shifted from the human side to deep learning algorithms in tasks regarding to image processing, natural language processing, and general gameplays. However, we have to admit that many existing deep learning algorithms strongly rely on manual data annotation for training.

Therefore, unsupervised learning has gradually attracted the attention of academia and industry in recent years, and has become a research hotspot in many fields. Taking natural language processing as an example, the achievements of unsupervised learning in this area are transformative. The word vectors trained by unsupervised methods such as Skip-Gram \cite{mikolov2013distributed}, GLoVe \cite{pennington2014glove}, Bert\cite{devlin2018bert} and many more have helped researchers to solve various downstream practical problems. Similarly, in the field of image processing, unsupervised learning is not uncommon. For example, RotNet \cite{gidaris2018unsupervised} tries to build a relationship between image rotation angle and its semantics; OPN \cite{lee2017unsupervised} leveraged the temporal relationships between images; AET \cite{zhang2019aet} explored the way of encoding transformations other than image data; other algorithms are developed from the perspective of clustering \cite{ji2019invariant, caron2018deep} and have demonstrated good performance in many downstream tasks. However, all these previous methods are dependent on the certain configurations of data transformation. A general unsupervised representation learning that fully leverages all kinds of data augmentation and transformation is still lacking.

The goal of our work is to learn deep convolutional neural network based semantic features in an unsupervised manner. Our proposed method follows the self-supervised paradigm and constructs the similarities between images as distance metrics in the embedding space by leveraging the inter-correlation between augmented variants of each training samples. The contributions of the proposed approach are three-fold:

\begin{itemize}
\item We proposed a novel self-supervised approach. The method is easy to use and simple to implement. It offers meaningful image features as demonstrated in the paper, and the performance of the method surpasses the state-of-the-art algorithms in our evaluations.
\item The features learned from our proposed approach can be transferred to different tasks. The approach shows great performance in unsupervised image classification throughout our experiments. Moreover, the method performs well enough even when the size of the training set is limited.
\item The proposed approach can be used to compare similarities between images, and is much faster than conventional image processing methods such as SIFT.
In the task of sorting similar pictures, many existing methods rely on pre-trained models based on supervised learning, while our method can completely relax this limitation, and only use unlabeled images as training data to obtain an even better performance compared to these methods, which thereby demonstrates that the performance gap between supervised and unsupervised representation learning is further narrowed.
\end{itemize}

In this paper, we introduce an unsupervised representation learning paradigm. The paper is organized as follows. First, we introduced our method in Section 2. Then the experimental results and training details are illustrated in Section 3. The conclusion and potential future works are discussed in the last section.

\section{Method}

\begin{figure}[!t]
    \centering
    \begin{subfigure}[b]{0.49\linewidth}
        \includegraphics[width=.99\linewidth]{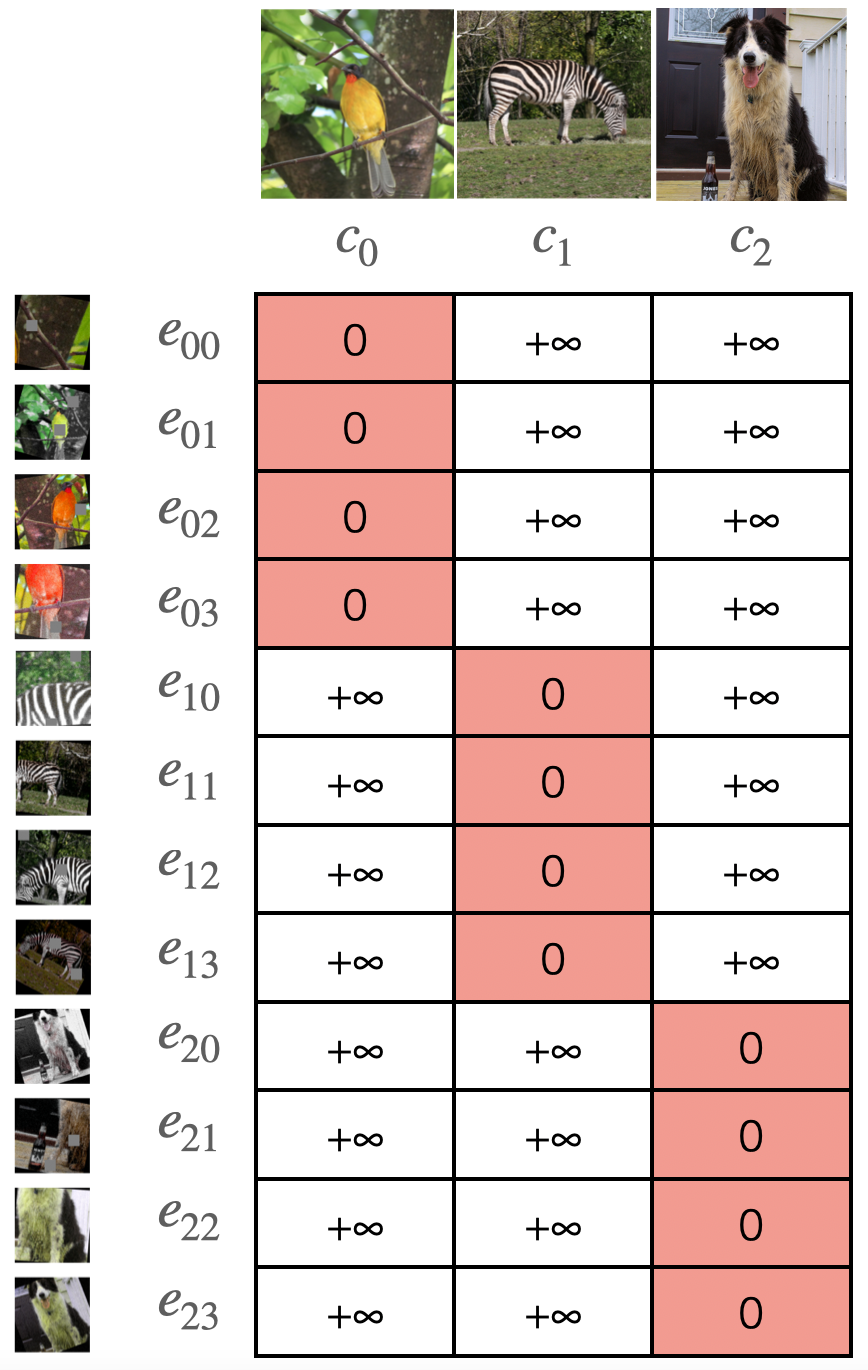}
        \caption{}
        \label{fig:loss_a} 
    \end{subfigure}
    \hfill
    \begin{subfigure}[b]{0.49\linewidth}
        \includegraphics[width=.99\linewidth]{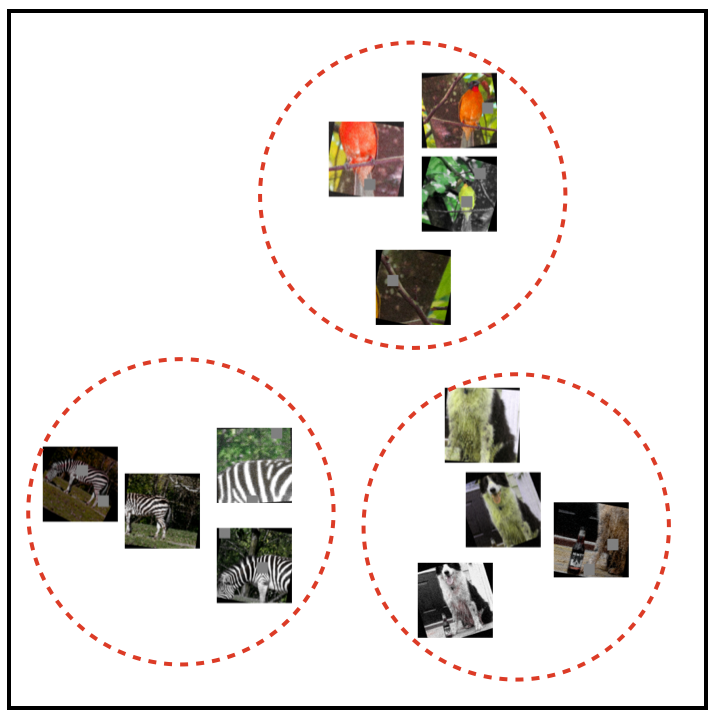}
        \caption{}
        \label{fig:loss_b} 
    \end{subfigure}
    \caption{Ground truth distance matrix (a) and an ideal clustering result (b).}\label{Fig:loss}
\end{figure}

\begin{figure*}
\centering

\begin{subfigure}[b]{.6\textwidth}
\centering
\includegraphics[width=\textwidth]{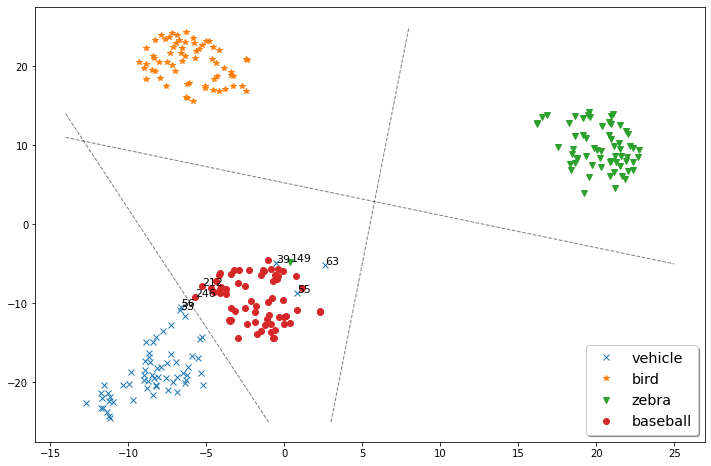}

\caption{}
\end{subfigure}\quad
\begin{subfigure}[b]{.25\textwidth}
\centering
\includegraphics[width=.45\textwidth]{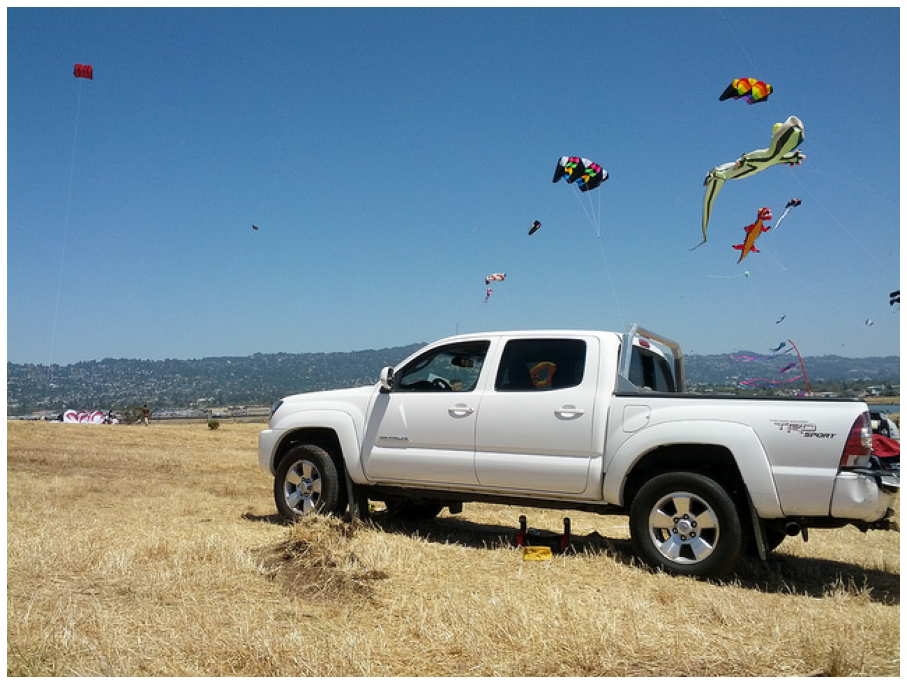}\quad
\includegraphics[width=.45\textwidth]{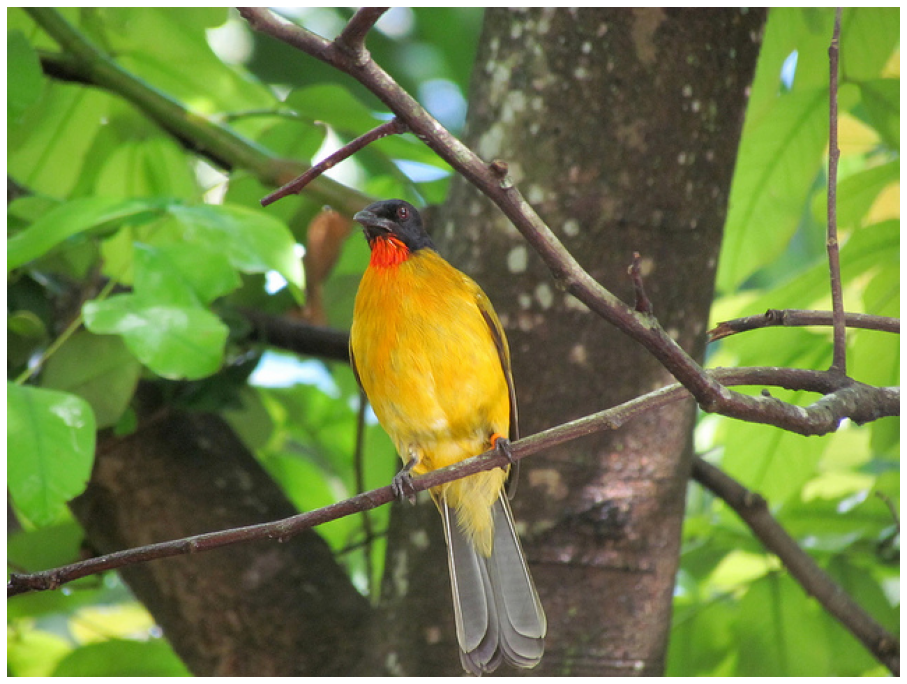}\quad
\includegraphics[width=.45\textwidth]{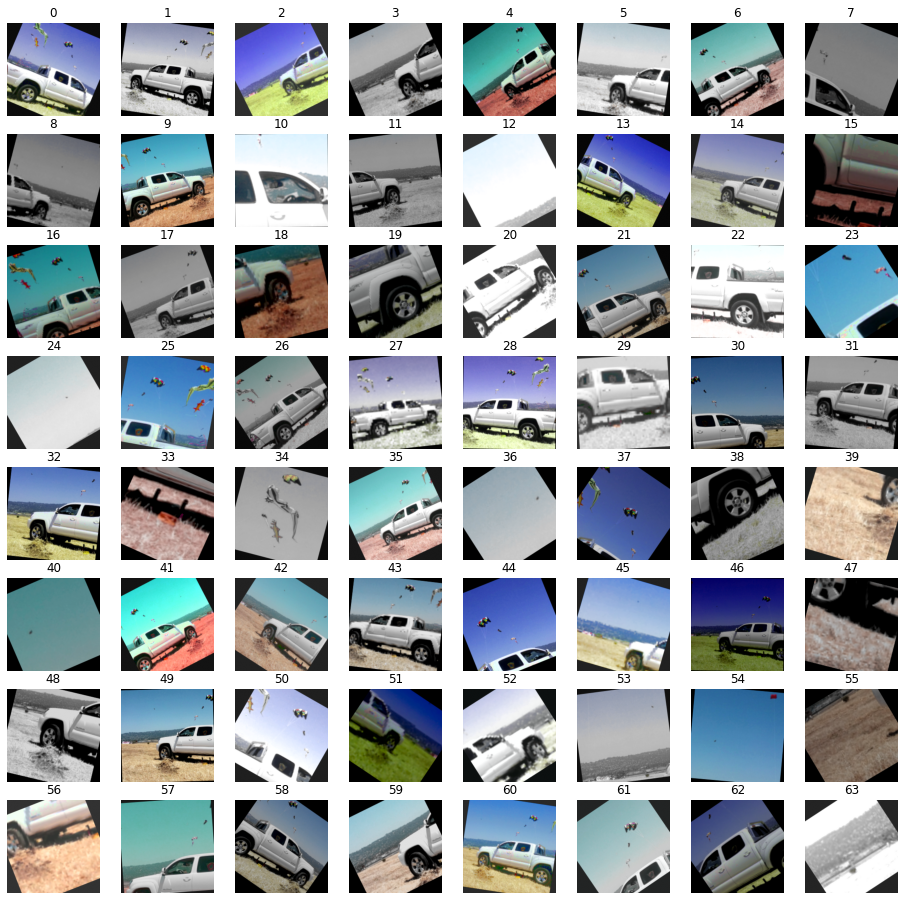}\quad
\includegraphics[width=.45\textwidth]{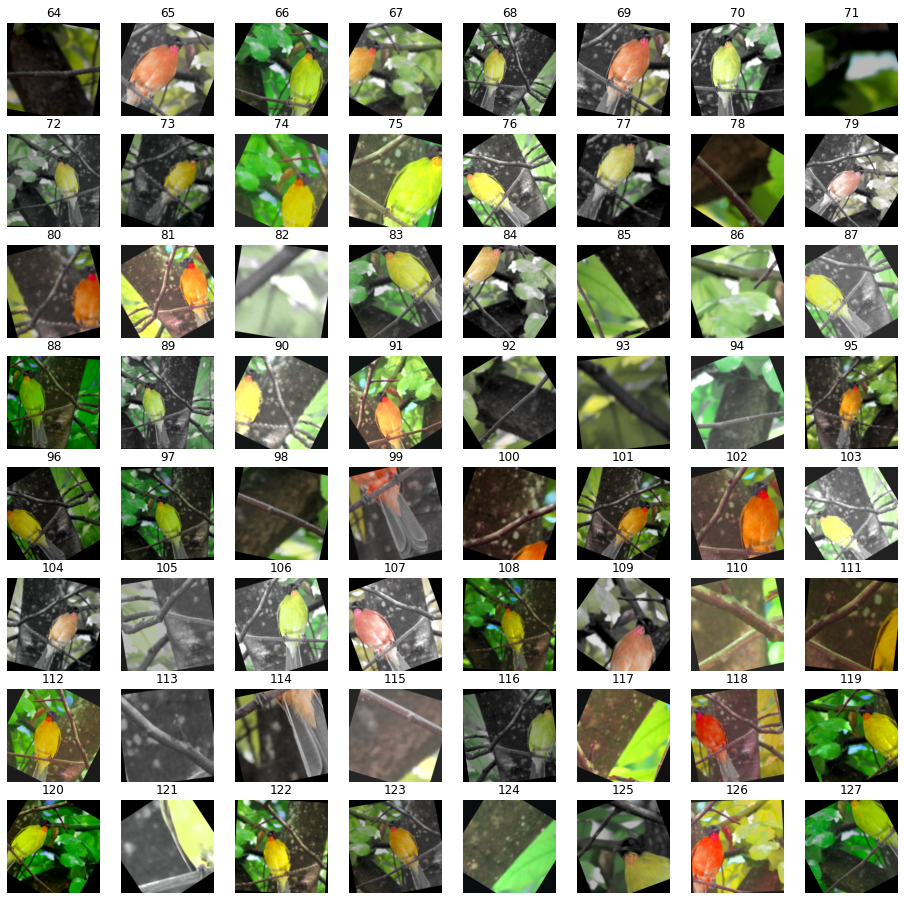}\quad
\includegraphics[width=.45\textwidth]{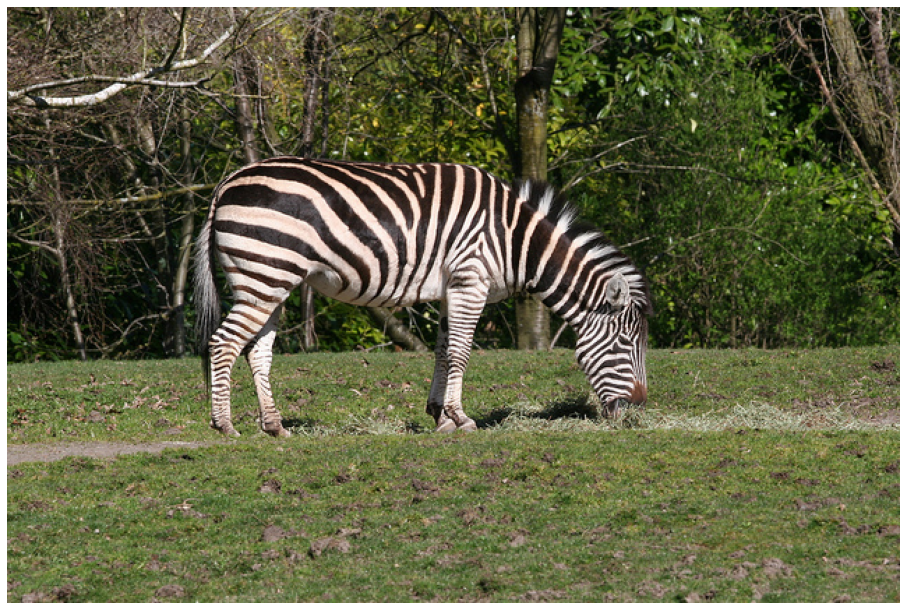}\quad
\includegraphics[width=.45\textwidth]{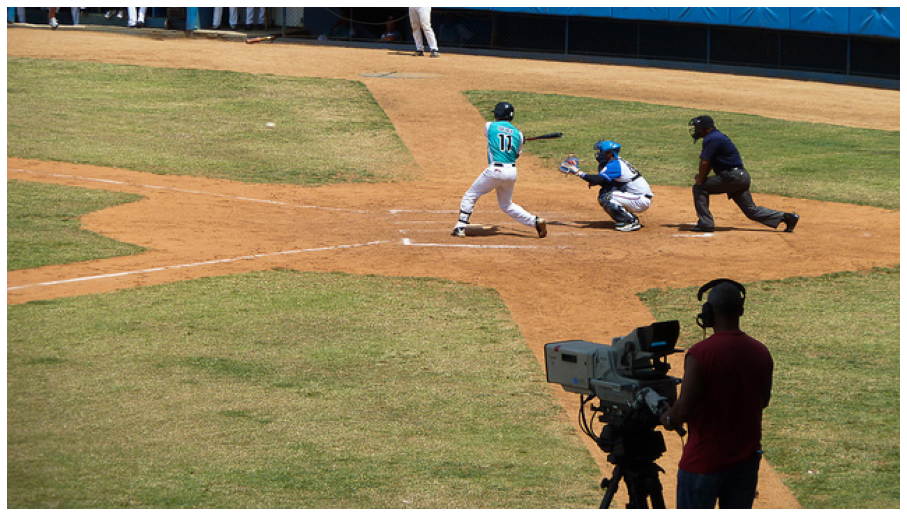}\quad
\includegraphics[width=.45\textwidth]{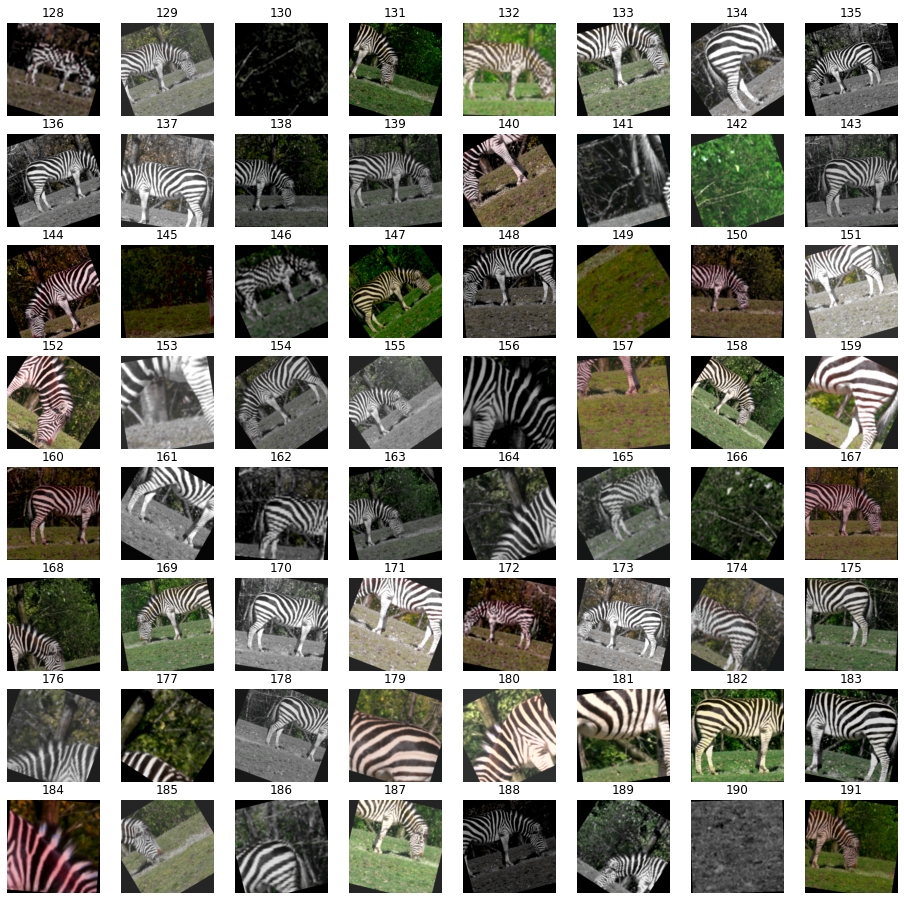}\quad
\includegraphics[width=.45\textwidth]{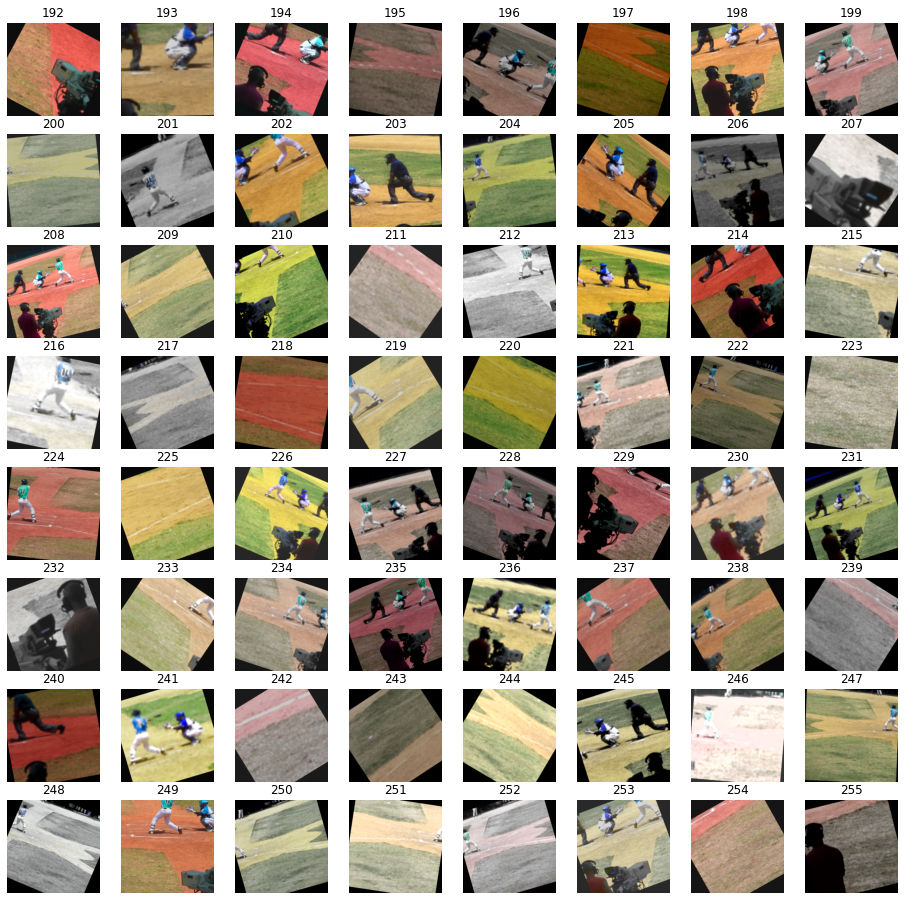}

\caption{}
\end{subfigure}
\begin{subfigure}[b]{.9\textwidth}
\centering
\includegraphics[width=\textwidth]{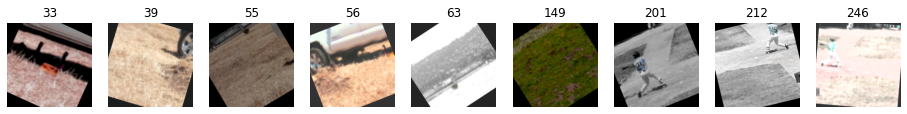}

\caption{}
\end{subfigure}

\caption{An illustration of Resnet50 Contrast (Table \ref{tab: 2}) clustering result. The four source pictures are marked with blue (vehicle), orange (bird), green (zebra), and red (baseball). (a) shows the position of each sampled picture's embedding in the two-dimensional coordinate system after t-SNE dimensionality reduction. (b) illustrates the source images and their augmented versions. We augmented and generated 64 small images from each source. (c) illustrates the outliers (39, 55, 63, and 149) and pictures locate on the border (33, 56, 201, 212, and 246).}\label{Fig:tsne}

\end{figure*}

The goal of the framework is to train a deep convolutional neural network (CNN) that extracts features from images in an unsupervised manner, and represents images as low-dimensional vectors while preserving the similarities between them. Our method is based on the intuition that the representations of images augmented from the same source picture should be close to each other in the emebdding space, while the distance between the representations should be far away from each other if the they are augmented from different sources.

We fetch $N \times M$ images to build a batch, which means the inputs are from $N$ source images and each is augmented to generate $M$ images. One image usually contains only a part of its source instead of the whole one. The purpose of this design is to enforce the neural network to effectively learn the inter-correlation between different parts of a same picture (such as the subject and the background), instead of just learning how the image is augmented. The embedding vector is defined as follow:

\begin{equation}
    \mathbf{e}_{ij} = \tanh(f(\mathbf{x}_{ij};w))\\
\end{equation}
Here we denote the output of the entire neural network as $f(\mathbf{x}_{ij};w)$ where $w$ represents all parameters of the neural network. $x_{ij}$ is the image data, and $\mathbf{e}_{ij}$ is the embedding vector of the $i^{th}$ image's $j^{th}$ augmentation.

Inspired by the idea of speaker verification \cite{wan2018generalized}, the desired distance matrix that stand for our purpose is illustrated in Figure \ref{fig:loss_a}, and the embedding vectors of images in the low-dimensional space which are generated by the CNN should be similar to what is shown in Figure \ref{fig:loss_b}. The loss function can be implemented in either of the two ways:

\textbf{Softmax} We apply a softmax function on the embeddings that classifies each augmented image into a source and the loss is defined using the cross entropy:

\begin{equation}
    \mathbf{p}_{ij,i} = \frac{\exp(-d(\mathbf{e}_{ij}, \mathbf{c}_{i}))}{\sum_{k=1}^{N}\exp(-d(\mathbf{e}_{ij}, \mathbf{c}_{k}))}\\
\end{equation}

\begin{equation}
\begin{split}
    \mathbf{L}_{ij} & = -\log(\mathbf{p}_{ij,i})\\
    & = d(\mathbf{e}_{ij}, \mathbf{c}_{i}) + \log\sum_{k=1}^{N}\exp(-d(\mathbf{e}_{ij}, \mathbf{c}_{k}))
\end{split}
\end{equation}
where

\begin{equation}
    \mathbf{c}_{i} = \frac{1}{M}\sum_{j=1}^{M}\mathbf{e}_{ij}\\
\end{equation}

\textbf{Contrast} We define the contrast loss with positive pairs and the most difficult negative pairs:

\begin{equation}
    \mathbf{L}_{ij} = d(\mathbf{e}_{ij}, \mathbf{c}_{i})-\min_{\substack{1<k<N \\ k\neq i}}d(\mathbf{e}_{ij}, \mathbf{c}_{k})\\
\end{equation}
In this approach, $d(\cdot,\cdot)$ is the L2 distance.

The augmentation methods appear in this paper include:

\begin{itemize}
\item \textbf{Rotation:} rotate the input image for a certain degree sampled from a uniform distribution $U(-35, 35)$.
\item \textbf{Noise:} add Gaussian noise from $N(0, 20)$ to the input image.
\item \textbf{Crop:} crop a square area from the original image and resize it to a specified size (e.g. $224 \times 224$). The ratio between the side length of the square and the length of the shorter side of the source image is sampled from a uniform distribution $U(0.2, 1)$. The lower limit of the distribution is denoted as the minimum crop rate $CR_{min}$ below.
\item \textbf{Resolution:} reduce the resolution of the image, and then resize it to its original size. The rate of resolution reduction is sampled from $U(0.1, 1)$.
\item \textbf{Hue:} increase or decrease the hue by a random value sampled from a uniformed distribution $U(-25, 25)$.
\item \textbf{Saturation:} increase or decrease the saturation by a random value sampled from a uniformed distribution $U(-150, 50)$. Since we hope the model to parse black-and-white photos, the distribution is not symmetric around zero.
\item \textbf{Brightness:} multiply the brightness by a random value sampled from a uniformed distribution $U(0.75, 1.25)$ and increases or decreases by a bias sampled from $U(-25, 25)$.
\item \textbf{Cutout \cite{devries2017improved}:} fill part of the input image with gray areas, where each area has 15\% of the corresponding size of the height and width. The number of areas is sampled from $U(0, 2)$.
\end{itemize}

In addition to the above augmentation methods, all augmented images have a 50\% probability of being flipped horizontally. 

Figure \ref{Fig:tsne} shows how a well trained network should behave. The images augmented from the same source is going to be clustered into the same pile, and the boundary between piles shall be as clear as possible. It can be seen from the figure that most of the outliers cannot be clustered correctly due to the lack of feature-rich textures. For example, when the entire canvas is filled with lawns, it is difficult to distinguish whether the augmented image comes from image 1 (vehicle), image 3 (zebra), or image 4 (baseball).

\section{Experiments}


In this section, we first explore the factors that affect the performance of our model. We compare the performance with other existing unsupervised models on multiple image classification and image retrieval tasks. If not specifically pointed out, all the following experiments are trained on 2 NVIDIA Tesla V100 (32GB) GPUs.

\subsection{Evaluations of training methods}

\begin{table}[!t]
    \centering
    \caption{Evaluation of different loss functions and the impact of network depth on results. The size of the input image is $32 \times 32$.}
    \begin{tabular}{l|cc|cc}
    \toprule
        \multirow{2}{*}{Model} & \multicolumn{2}{|c|}{COCO 2014} & \multicolumn{2}{c}{Imagenet} \\
        & Top 1 & Top 10 & Top 1 & Top 10 \\
        \toprule
        3 Conv. (Softmax) & 13.09 & 43.36 & 14.45 & 43.36 \\
        3 Conv. (Contrast) & \textbf{94.63} & \textbf{99.02} & 92.48 & \textbf{98.63} \\
        4 Conv. (Contrast) & 91.60 & 98.04 & 90.63 & 98.14 \\
        5 Conv. (Contrast) & 92.09 & \textbf{99.02} & \textbf{92.58} & \textbf{98.63} \\
    \toprule
    \end{tabular}
    \label{tab: 1}
\end{table}

\begin{table}[!t]
    \centering
    \caption{Evaluation of different loss functions and the impact of network depth on results. The size of the input image is $224 \times 224$.}
    \begin{tabular}{l|cc|cc}
    \toprule
        \multirow{2}{*}{Model} & \multicolumn{2}{|c|}{COCO 2014} & \multicolumn{2}{c}{Imagenet} \\
        & Top 1 & Top 10 & Top 1 & Top 10 \\
        \toprule
        Res18 (Contrast) & 73.54 & 85.25 & 72.75 & 84.86 \\
        \hline
        Res50 (Softmax) & 19.62 & 47.16 & 27.93 & 59.86 \\
        Res50 (Contrast) & \textbf{85.64} & \textbf{95.70} & \textbf{86.82} & \textbf{97.27} \\
    \toprule
    \end{tabular}
    \label{tab: 2}
\end{table}

We start by comparing the two loss functions described in the previous section. In the first experiment, the ImageNet training dataset is used as the source images. The Network in Network (NIN) \cite{lin2013network} architecture is implemented as the backbone. The implementation detail follows the model built by Gidaris \etal \cite{gidaris2018unsupervised}, where a Conv. block in the NIN architectures have 3 convolutional layers. The minimum crop rate $CR_{min}$ is set to 0.5. Each mini-batch contains 1024 images (N=1024) and each image is augmented 8 times (M=8). The images are resized to $32 \times 32$. The models are trained for 20 hours using the Adam optimizer where the initial learning rate is $5 \times 10^{-4}$.

During the evaluation, both ImageNet's validation set as well as COCO 2014's validation set are used to ensure the generalization ability and the models are not overfitted. The first 512 images from each dataset are picked. By augmenting each image once (all of the above augmentation operations will be executed except for cropping), a total of 1024 images are obtained. Then a square area with the same side length as the short side is cutted out from each image. This operation makes the contents displayed by the two pictures in each picture pair are not exactly the same. The models encode each image into a 192-dimension vector. Then it is able to find the $k$ nearest neighbours to each picture, and verify whether the corresponding original or augmented picture to an image is in its $k$ neighboring pictures. The results in Table \ref{tab: 1} show that the contrast loss outperforms the softmax loss drastically. We also perform tests on relatively high-definition pictures (where the size of images increase to $224 \times 224$) with a more complex network, and the test results in Table \ref{tab: 2} also demonstrate the strength of contrast loss compared to softmax loss. To train the networks, each mini-batch contains 64 images (N=64) and the images are augmented 8 times (M=8).

\subsection{Evaluations of feature quality}

\begin{table}[!t]
\caption{Feature quality evaluation of feature maps with different depths. The linear classifier maps the outputs to 10 classes. The non-linear classifier contains 2 hidden layers have 256 feature channels each.}
\resizebox{\linewidth}{!}{%
    \centering
    \begin{tabular}{l|cccccc}
    \toprule
        Model & Conv1 & Conv2 & Conv3 & Conv4 & Conv5 & Output\\
        \toprule
        \multicolumn{7}{l}{\textbf{Random Initialization + Linear Classifiers}} \\
        \hline
        5 Conv. Blocks & 57.38 & 33.53 & 15.08 & 10.25 & 10.00 & 10.01 \\
        \hline
        \multicolumn{7}{l}{\textbf{Linear Classifiers}} \\
        \hline
        3 Conv. Blocks & 72.07 & 72.97 & 69.21 & - & - & 56.73 \\
        4 Conv. Blocks & 72.32 & 72.54 & 74.57 & 69.58 & - & 57.78 \\
        5 Conv. Blocks & 72.48 & 73.17 & \textbf{74.78} & 74.09 & 68.59 & 57.37 \\
        \hline
        \multicolumn{7}{l}{\textbf{Non-Linear Classifiers}} \\
        \hline
        3 Conv. Blocks & 78.39 & 79.96 & 75.45 & - & - & 65.21 \\
        4 Conv. Blocks & 78.54 & 80.64 & 78.77 & 75.15 & - & 66.36 \\
        5 Conv. Blocks & 78.55 & \textbf{80.72} & 79.18 & 76.39 & 74.32 & 66.98 \\
    \toprule
    \end{tabular}}
    \label{tab: layers}
\end{table}

\begin{table}[!t]
\caption{Evaluation of different augmentation methods. The reported results are from CIFAR-10. For all the entries we train a linear classifier on top of the feature maps generated by the 3rd conv. block of a AugNet model with 5 conv. blocks in total.}
\resizebox{\linewidth}{!}{%
    \centering
    \begin{tabular}{ccccccc|c}
    \toprule
        \multicolumn{7}{c|}{Augmentation} & \multirow{2}{*}{Accuracy}\\
        \cline{1-7}
        Rotation & Noise & Crop & Hue & Saturation & Brightness & Cutout & \\
        \toprule
        X & X & X & X & X & X & X & 15.08 \\
        \hline
        $\bigcirc$ & X & X & X & X & X & X & 61.04 \\
        \hline
        $\bigcirc$ & $\bigcirc$ & X & X & X & X & X & 60.29 \\
        $\bigcirc$ & X & $\bigcirc$ & X & X & X & X & 64.13 \\
        $\bigcirc$ & $\bigcirc$ & $\bigcirc$ & X & X & X & X & 63.78 \\
        \hline
        $\bigcirc$ & $\bigcirc$ & $\bigcirc$ & $\bigcirc$ & X & X & X & 67.87 \\
        $\bigcirc$ & $\bigcirc$ & $\bigcirc$ & X & $\bigcirc$ & X & X & 72.27 \\
        $\bigcirc$ & $\bigcirc$ & $\bigcirc$ & X & X & $\bigcirc$ & X & 71.03 \\
        $\bigcirc$ & $\bigcirc$ & $\bigcirc$ & $\bigcirc$ & $\bigcirc$ & $\bigcirc$ & X & 76.80 \\
        \hline
        $\bigcirc$ & $\bigcirc$ & $\bigcirc$ & $\bigcirc$ & $\bigcirc$ & $\bigcirc$ & $\bigcirc$ & 74.78 \\
    \toprule
    \end{tabular}}
    \label{tab: augment}
\end{table}

We then evaluate the feature quality of models' intermediate outputs. The three versions of AugNets with contrast loss in Table \ref{tab: 1} are implemented in this section. We train linear and non-linear classifiers on top of the feature maps generated by the convolutional blocks as well as the final outputs. The linear classifier simply maps the outputs to 10 classes, while the non-linear classifier contains 2 hidden layers with each containing 256 feature channels followed by batch-norm and ReLU units. The final experimental results in Table \ref{tab: layers} are very similar to those proposed by a previous study \cite{gidaris2018unsupervised}, which shows that as the depth increases, the outputs may be more specialized in the designed tasks. This again proved that the intermediate layers has the potential for better semi-supervised learning. The backbones' training configurations are identical to the experiments shown in Table \ref{tab: 1}, and the linear/non-linear classifiers are trained on one Nvidia Tesla V100 (32GB) GPU for 200 epochs.

In Table \ref{tab: augment}, we further test the impact of different augmentation methods. For this purpose, we design 10 sets of augmentation strategies, and the details of each augmentation method is described in the Method section. We use the feature maps generated by the $3^{rd}$ conv. block of an AugNet model with 5 conv. blocks together with a linear classifier (which shows the best performance in Table \ref{tab: layers}). It can be observed that most augmentation methods can effectively improve the accuracy of the classifier. This may be because more diverse augmentation schemes allow the model to better learn the inter-correlation between local features at different positions in a picture, rather than learning some speculative strategies which can lead to overfitting.

\begin{table}[!t]
\caption{The clustering performance on three object image benchmarks. We report classification accuracy averaged over 5 experiments. Numbers for other methods are from the original papers.}
\resizebox{\linewidth}{!}{%
    \centering
    \begin{tabular}{l|ccc}
    \toprule
        
        Model & STL10 & CIFAR10 & CIFAR100\\
        \toprule
        \multicolumn{2}{l}{\textbf{Supervised Methods}} \\
        \hline
        NIN \cite{lin2013network} & - & 91.19 & 64.32* \\
        \hline
        \multicolumn{2}{l}{\textbf{Unsupervised Methods}} \\
        \hline
        Random network & 13.5 & 13.1 & 5.93 \\
        K-means \cite{NIPS2004_40173ea4} & 19.2 & 22.9 & 13.0 \\
        SC \cite{wang2014optimized} &  15.9 & 24.7 & 13.6 \\
        AE \cite{bengio2007greedy} & 30.3 & 31.4 & 16.5 \\
        GAN \cite{radford2015unsupervised} & 29.8 & 31.5 & 15.1 \\
        JULE \cite{yangCVPR2016joint} &  27.7 & 27.2 & 13.7 \\
        DEC \cite{xie2015Unsupervised} & 35.9 & 30.1 & 18.5 \\
        DAC \cite{Chang_2017_ICCV} &  47.0 & 52.2 & 23.8 \\
        DeepCluster \cite{caron2018deep} & 33.4 & 37.4 & 18.9 \\
        ADC \cite{haeusser2019Associative} & 53.0 & 32.5 & 16.0 \\
        IIC \cite{ji2019invariant} & 59.6 & \textbf{61.7} & 25.7 \\
        AugNet ($Avg \pm Std$) & $58.60 \textcolor{blue}{\pm 0.85}$ & $56.21 \textcolor{blue}{\pm 2.11}$ & $27.68 \textcolor{blue}{\pm 0.40}$ \\
        AugNet ($Best$) & \textbf{60.25} & 59.88 & \textbf{29.19} \\
        AugNet+5k ($Avg \pm Std$) & $52.97 \textcolor{blue}{\pm 1.24}$ & $34.19 \textcolor{blue}{\pm 0.94}$ & $19.72 \textcolor{blue}{\pm 0.69}$ \\
        AugNet+5k ($Best$) & 54.58 & 35.88 & 20.59 \\
        
    \toprule
    \end{tabular}}
    \label{tab: cifar}
\end{table}

\begin{table*}[!t]
\caption{Evaluation of image retrieval quality. We report mAP averaged over 5 experiments. Numbers for other methods are mostly obtained using our experiments with authors’ original code. *: The result is obtained using our experiment with OpenCV's implementation, where images are resized to $224 \times 224$. **: Results are from the original papers.
}
    \centering
    \begin{tabular}{l|cccc}
    \toprule
        \multirow{2}{*}{Model} & \multicolumn{4}{c}{mAP} \\
        \cline{2-5}
        & Paris6k & Pok\'emon & Anime Face & Human Sketch \\
        \toprule
        SIFT & 17.6* & 57.32 & 6.26 & 3.00 \\
        \hline
        \multicolumn{5}{l}{\textbf{Visual Representation Learning}} \\
        \hline
        Doersch \etal \cite{doersch2015unsupervised} & 53.1** & & & \\
        Wang \etal \cite{wang2017transitive} & 58.0** & & & \\
        \hline
        \multicolumn{5}{l}{\textbf{Image Retrieval Methods}} \\
        \hline
        VGG16-GeM \cite{2016CNN} & 87.8** & 73.63 & 11.40 & 15.24 \\
        ResNet101-GeM \cite{2016CNN} & 92.5** & 68.77 & 12.29 & 15.29 \\
        GL18-TL-ResNet-50-GeM-W \cite{2016CNN, radenovic2018fine} & 91.64 & 77.45 & 16.70 & 16.99 \\
        GL18-TL-ResNet-101-GeM-W \cite{2016CNN, radenovic2018fine} & 93.09 & 78.38 & 13.79 & 17.52 \\
        GL18-TL-ResNet-152-GeM-W \cite{2016CNN, radenovic2018fine} & 92.95 & 80.58 & 15.04 & 16.65 \\
        rSfM120k-TL-ResNet-50-GeM-W \cite{2016CNN, radenovic2018fine} & 90.23 & 79.56 & 17.44 & 17.67 \\
        rSfM120k-TL-ResNet-101-GeM-W \cite{2016CNN, radenovic2018fine} & 91.66 & 77.39 & 15.37 & 19.20 \\
        rSfM120k-TL-ResNet-152-GeM-W \cite{2016CNN, radenovic2018fine} & 91.45 & 80.44 & 15.44 & 18.79 \\
        Resnet101-TL-MAC \cite{gordo2017end, revaud2019learning} & 90.1** & 73.31 & 11.10 & 6.20 \\
        Resnet101-TL-GeM \cite{gordo2017end, revaud2019learning} & \textbf{93.4}** & 73.30 & 10.77 & 7.03 \\
        Resnet50-AP-GeM \cite{gordo2017end, revaud2019learning} & 91.9** & 68.85 & 8.56 & 7.20 \\
        Resnet101-AP-GeM \cite{gordo2017end, revaud2019learning} & 93.0** & 69.75 & 10.79 & 6.35 \\
        Resnet101-AP-GeM-GL18 \cite{gordo2017end, revaud2019learning} & 93.1** & 72.07 & 9.20 & 7.07 \\
        \hline
        AugNet ($Avg \pm Std$) & $60.30 \textcolor{blue}{\pm 1.11}$ & $85.02 \textcolor{blue}{\pm 1.40}$ & $20.10 \textcolor{blue}{\pm 0.14}$ & $19.25 \textcolor{blue}{\pm 0.57}$ \\
        AugNet ($Best$) & 61.18 & \textbf{86.50} & \textbf{20.33} & \textbf{20.03} \\
    \toprule
    \end{tabular}
    \label{tab: siamac}
\end{table*}

We evaluate our method on image classification task with STL-10, CIFAR-10, and CIFAR-100 datasets. From Table \ref{tab: cifar}, we can see that our method exhibits competitive performance and achieves a better record than the existing state-of-the-art method in some tasks. In the above model, we use Resnet18 as the backbone and STL-10 unlabeled dataset (containing 100,000 images) as the training set. The images are all subjected to the same augmentation pipeline (excluding Gaussian noise and cutout) before being fed into the network. The augmented pictures are all resized to a square with side length equal to 224 to facilitate neural network processing while training. As for the CIFAR-10 and CIFAR-100 tasks, their training sets (each contains 50,000 pictures) are also added as the training data, respectively. The sizes of the training sets used for these three datasets of STL-10, CIFAR-10, and CIFAR-100 are 100K, 150K, and 150K, respectively. K-Means is used for clustering embeddings. It is worth noting that, even if we use less training data, this method can still achieve relatively good results, even surpassing most classic methods as shown in the last two rows of Table \ref{tab: cifar}. In this test, the STL-10 training set (contains 5000 images) is the only training source, and the backbone is Resnet18. Since the average cropping rate is 0.6, to keep the receptive field of each convolution kernel consistent, the input images are resized so that the lengths of shorter sides during inference equals 370.


\subsection{Image retrieval}

\begin{figure}[!b]
    \centering
    \includegraphics[width=.99\linewidth]{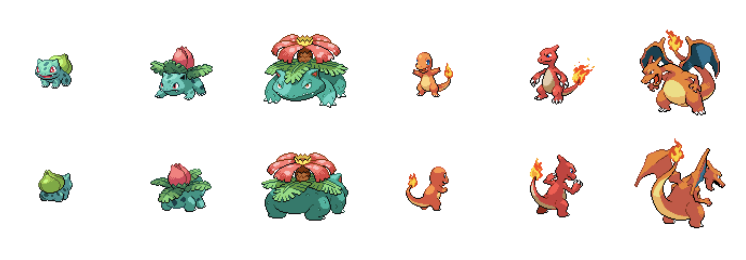}
    \caption{Front and back pictures of the first 6 Pok\'emons.}\label{Fig:cartoon_example_pokemon}
\end{figure}

\begin{figure*}[!t]
    \centering
    \includegraphics[width=.99\linewidth]{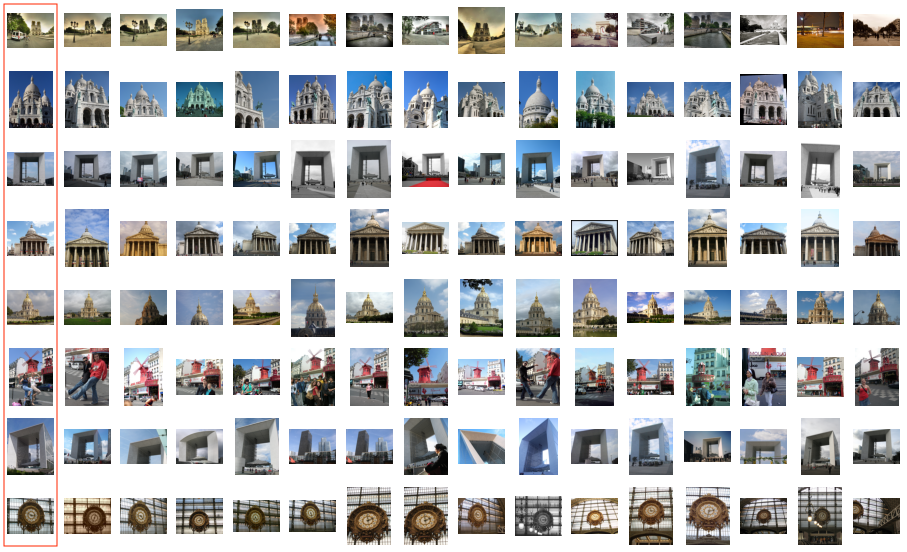}
    \caption{Image retrieval result for the Paris6k datasets. The left most images are the queries.}\label{Fig:sim_find_vis_wide2}
\end{figure*}

\begin{figure*}[!t]
    \centering
    \begin{subfigure}[t]{0.99\linewidth}
        \includegraphics[width=\linewidth]{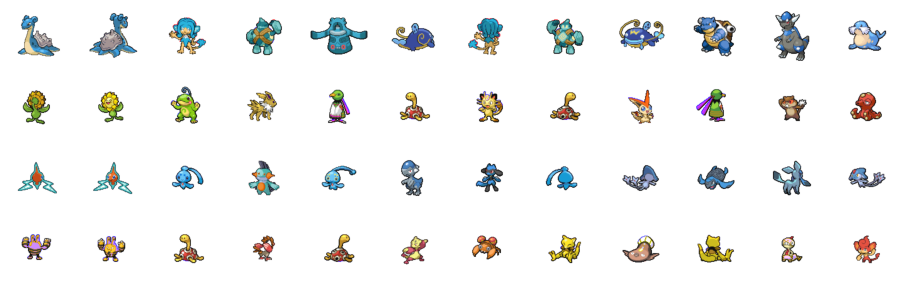}
        \caption{Pok\'emon}\label{fig:cartoon_pokemon}
    \end{subfigure}
    \begin{subfigure}[t]{0.99\linewidth}
        \includegraphics[width=\linewidth]{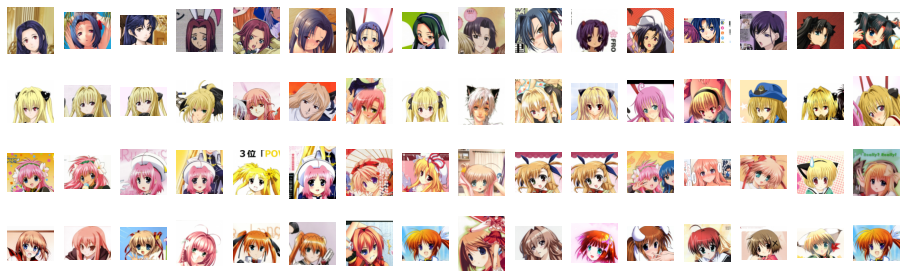}
        \caption{Tagged Anime Illustrations}\label{fig:cartoon_faces}
    \end{subfigure}
    \begin{subfigure}[t]{0.99\linewidth}
        \includegraphics[width=\linewidth]{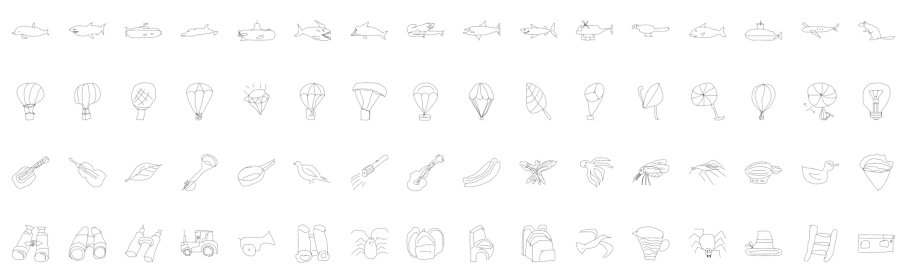}
        \caption{Humans Sketchs}\label{fig:cartoon_sketch}
    \end{subfigure}
    \caption{Image retrieval result for out-of-domain datasets. The left most images are the queries.}\label{Fig:cartoons}
\end{figure*}

Since our method borrows some ideas of clustering and recognition in principle, we evaluate the performance of this algorithm on some image retrieval tasks in Table \ref{tab: siamac}. Although the accuracy is slightly higher than some other methods based on visual representation learning, our method has almost no competitiveness compared with many well-designed image retrieval algorithms on standard testsets taking the Paris6k \cite{philbin2008lost} dataset as an example. However, the above-mentioned discrepancy is mainly due to the differences in training purpose and training process. Generally speaking, if one want to achieve better image retrieval results on a standard dataset, a pre-trained model based on the same or a similar data distribution is needed. On the basis of this model, the algorithm also needs to perform a complex comparison on its feature maps, instead of simply comparing the Euclidean distance of two vectors. Although these methods can achieve amazing results on standard datasets, in practice, we cannot find a suitable pre-training dataset in all situations. Hence, we assume that when the distribution of the pre-training dataset is far from the inference data's, the performance will be lower than expectation due to the lack of generalization. Based on the this assumption, we prepare three out-of-domain datasets:

\textbf{Pok\'emon} The Pok\'emon dataset \cite{veekun} contains images of 648 Pok\'emons. Each Pok\'emon has two pictures taken from the front and back as shown in Figure \ref{Fig:cartoon_example_pokemon}, so that there are 1296 images in the dataset, and the number of queries is also 1296. The ground truth of each query is the front/back picture corresponding to itself.

\textbf{Tagged Anime Illustrations} The Tagged Anime Illustrations dataset \cite{taggedanime} contains more than 14,000 face pictures over 173 anime characters. We use a subset for the purpose of evaluation, where each character has 30 pictures, and only one of them is in the query set. Our subset contains 5190 images in total, and 173 queries.

\textbf{Humans Sketchs} The dataset \cite{eitz2012hdhso} contains 20,000 unique human sketches evenly distributed over 250 object categories. We also use a subset for the purpose of evaluation, and each category contains 30 pictures, where one of them is in the query set. The subset contains 7500 images in total, and 250 queries.

From the results shown in the Table \ref{tab: siamac}, we report that our algorithm outperforms all other image retrieval methods based on pre-trained models. Our self-supervised model achieves is 86.50\%, 20.33\%, and 20.03\% mAP on the Pok\'emon dataset, the Tagged Anime Illustration dataset, and the Human Sketches dataset, respectively, where each is 5.92, 3.63, and 0.83 percentage points higher than the previous state-of-the-art methods.

In Figure \ref{Fig:sim_find_vis_wide2}, we can see that despite the fact that our evaluation results on the Paris6k dataset have some gaps with other supervised image retrieval methods, in general, our method gives an acceptable similarity measurement result. In the figure, it is apparent that the top 15 similar pictures of each query are correct and the style of the shots are analogous to the corresponding queries. In Figure \ref{Fig:cartoons}, we illustrate the performance of our method on the three out-of-domain datasets. As we mentioned above, even if the dataset is small, this method can achieve good results. For example in Figure \ref{fig:cartoon_pokemon}, the Pok\'emon dataset only contains no more than 1,300 pictures, but the model can still accurately find the corresponding image to each query, while the rest of the similar pictures found also have closing styles, color schemes, or analogous body parts to the query. However, it is not difficult to find from Table \ref{tab: siamac} that neither our model nor other supervised algorithms can achieve such good results on the anime character avatar or sketch dataset as on Paris6k. Since some Japanese anime characters tend to be homogenized in their creation, such as a relatively large head, a pointed chin, and big eyes, it is more difficult to find pictures of the same character from a large amount of avatar images. As for the human sketch dataset, we conclude that the main reason for this difference is people are usually more casual when drawing sketches, which make it difficult for the model to learn the inter-correlation between the various parts of the images.

We use Resnet18 as the backbone for training the Pok\'emon dataset, and Resnet34 for the remaining two datasets. When training the Human Sketches dataset, we take the average of channels and convert the pictures of three channels into single-channel images. We train 8, 16, and 32 epochs respectively on these three datasets, and they each took about 20 minutes, 6 hours, and 12 hours to converge. The images are resized so that the shorter sides' lengths equal 370  during inference.

\section{Conclusions}

In this paper, we propose a novel and straightforward unsupervised end-to-end visual representation learning method that can extract high-fidelity features from images which can be used as inputs for various downstream tasks such as image classification and image retrieval. We analyzed the performance of the model both internally and externally. From the perspective of the model itself, we compare the efficiency of two loss functions that both can realize our ideas. we show that our model is the most suitable and best performing among all models in various applications. Our future works include evaluating the image representation learned from our model on other common downstream tasks such as object detection and image segmentation, and extending the ideas underlying our model to other modalities of data such as videos and point clouds.

\section{Acknowledgements}

We are deeply grateful to Ziyi Yang for his insightful comments and many useful suggestions. We sincerely wish him every success in the final period of his Ph.D. study.

{\small
\bibliographystyle{egbib}
\bibliography{egbib}
}

\end{document}